\RequirePackage{fix-cm}
\documentclass[smallextended]{svjour3}       
\smartqed  
\usepackage{bbding}
\usepackage{algorithm}
\usepackage{algorithmic}
\usepackage{amssymb}
\usepackage{color}
\usepackage{graphicx}
\usepackage{subfigure}
\usepackage{amsmath}
\usepackage{subfigure}
\usepackage{hhline}
\begin{document}

\title{Applying Ricci Flow to High Dimensional Manifold Learning
}


\author{Yangyang Li$^{1,2}$        \and
         Ruqian Lu$^1$ 
}


\institute{\Envelope \quad Yangyang Li \at
              $^1$ Academy of Mathematics and Systems Science Key Lab of MADIS
              Chinese Academy of Sciences, Beijing \at
              $^2$ University of Chinese Academy of Sciences, Beijing\\
              \email{liyangyang12@mails.ucas.ac.cn}           
           \and
}

\date{Received: date / Accepted: date}

\maketitle

\begin{abstract}
In machine learning, a high dimensional data set such as the digital image of human face is often considered as a point set distributed on a (differentiable) manifold. In many cases the intrinsic dimension of this manifold is low but the representation dimension of data points is high. In order to ease data processing, one uses manifold learning (ML) techniques to reduce a high dimensional manifold (HDM) to a low dimensional one while keeping its essential geometric properties such as relative distances between points unchanged. Traditional ML algorithms often bear an assumption that the local neighborhood of any point on HDM is roughly equal to the tangent space at that point. This assumption leads to the disadvantage that the neighborhoods of points on the manifold, though with very different curvature, will be treated equally and projected to lower dimensional space in the same way. The curvature indifferent way of manifold processing often makes traditional dimension reduction poorly neighborhood preserving. To overcome this drawback we propose to perform an ‘operation’ on the HDM with help of Ricci flow before manifold’s dimension reduction. More precisely, with Ricci Flow we transform each local neighborhood of the HDM to a constant curvature patch. The HDM as a whole is then transformed to a subset of a Sphere with constant positive curvature. We compare our algorithm with other traditional manifold learning algorithms. Experimental results have shown that our method outperforms other ML algorithms with better neighborhood preserving rate.

\keywords{Manifold Learning \and Ricci flow \and Ricci Curvature \and Dimension Reduction}
\end{abstract}

\section{Introduction}
\label{intro}
In many machine learning tasks, one is often confronted with redundant dimensions of data representation. Manifolds usually arise from data generated in some continuous process. The generated manifold is often embedded in some high-dimensional Euclidean space. In most of the cases the manifold is represented as a discrete data set. An intuitive example is a set of images generated by a continuous changing of face expressions \cite{37}. Actually, this set of data points can be represented by just a low dimensional set of features accurately. In order to uncover these features, S. Roweis et al. \cite{13} and J. Tenenbaum et al. \cite{14} introduced a new research field called manifold learning (in this paper, manifold learning only refers to nonlinear dimensionality reduction technique). It is a perfect combination of classical geometric analysis and computer science. Traditional manifold learning algorithms aim to reduce manifold's dimension, so that its lower dimensional representation (i.e. the features) could reflect the intrinsic geometric and topologic structure of the high dimensional data points. In general, the existing manifold learning algorithms can be roughly divided into two classes: one is to preserve the global geometric structure of manifold, such as Isomap \cite{14}; the other one is to preserve the local neighborhood geometric structure, such as LLE \cite{13}, LEP \cite{15}, LPP \cite{18}, LTSA \cite{17}, Hessian Eigenmap \cite{16} et al. Isomap aimed to preserve the geodesic distance between any two high dimensional data points, which can be viewed as an extension of Multidimensional Scaling (MDS) \cite{30}. Local neighborhood preserving algorithms approximated manifolds with a union of locally linear patches (possibly overlapped). After the local patches are estimated with linear methods such as PCA \cite{19}, the global representation is obtained by aligning the local patches together. Besides using geometry tools, a lot of manifold learning algorithms used statistical knowledge to uncover the lower dimensional structure of manifold \cite{36}. Manifold learning algorithms have been applied to many fields such as data dimension compression, computer vision, image recognition and more. Despite the success of manifold learning, there are still several problems remaining. 

\begin{itemize}
	\item  \textbf{Locally short circuit problem:} if the embedded manifold is highly curved, the Euclidean distance between any two points is obviously shorter than the intrinsic geodesic distance. In this case the geodesic distance between these two points is often underestimated. 
	\item  \textbf{Intrinsic dimension estimation problem:} since tangent spaces are simply taken as local patches, the intrinsic dimension of manifold cannot be determined by the latter, in particular in case of strongly varying curvature. 
	\item  \textbf{Curvature sensitivity problem:} if the curvature of original manifold is especially high at some point, quite many patches are needed for representing the neighborhoods around this point. But one may not have as many data points as needed to produce enough patches, especially when the data points are sparse. 
\end{itemize} 

Among the above mentioned three problems, the third one is critical. It is the background of the other two. To solve this problem is the main target of this paper.

\subsection{Motivation}

In manifold learning one usually maps an irregularly curved manifold to a lower dimensional Euclidean space directly. This is often unpractical since care is not taken of the varying geometric structure of the manifold at different points. Ricci flow (also called intrinsic curvature flow) is a very useful tool for evolving an irregular manifold and making it converging to a regular one (constant curvature manifold). It has been used to prove the \textsl{Poincare conjecture} \cite{31}. In this paper we use Ricci flow to regularize the metric and curvature of the generated manifold before reducing its dimension. Our algorithm RF-ML transforms the irregular manifold to a regular constant curvature manifold. Since Ricci flow preserves local structure of manifold and the Riemannian metric of the manifold after Ricci flow is uniform (see section 2), the local relationships among data points in the whole process of algorithm are preserved. The key thought of our method is to construct a Ricci flow equation directly on discrete data points without parametrization and meshes. Under the premise of non-negative curvature, the Ricci flow process evolves the original irregular manifold to a subset of a sphere. Finally traditional manifold learning algorithms can be used to reduce the dimensionality of the (high dimensional) sphere. It is assumed that the data points are distributed on a single open Riemannian manifold to guarantee that the manifold evolves into a punctured sphere under Ricci flow. In the current presentation we only consider Riemannian manifolds with non-negative Ricci curvature. Manifold learning with negative Ricci curvature will be discussed in our next paper. We call this style of algorithm dynamic manifold learning. This paper is just a first attempt in this direction. 

\subsection{Related Work}

In recent years, there have been several works studying intrinsic curvature of sampled points by combining traditional manifold learning techniques in designing curvature aware manifold learning algorithms. Among them Kwang et al. \cite{3} studied this problem with redefined group Laplacian operator. The redefined group Laplacian matrix was used to measure the pair-wise similarities between data points regarding point to point distance as well as curvature of the local patches. Their experimental results showed that the redefined Laplacian operator led to lower error rates in spectral clustering than the traditional Laplacian operator. W. Xu et al. \cite{4} \cite{5} \cite{6} used Ricci Flow to rectify the pair-wise non-Euclidean dissimilarities among data points and to combine traditional manifold learning algorithms to come up with a new dimension reduction algorithm. Another nonlinear dimension reduction algorithm is the two dimensional Discrete Surface Ricci Flow \cite{1}, which mainly applied to three dimensional data points distributed on a two dimensional manifold. Chow and Luo have studied the relations between circle packing metric and surface Ricci flow in a theoretical view. Gu et al. \cite{1} \cite{2} constructed discrete triangle meshes on three dimensional data points as well as discrete circle packing metric on the triangle meshes. They proposed the discrete surface Ricci flow algorithm using Newton’s method, which can map any curved surface to a normal two-dimensional surface (Sphere, Hyperbolic space, or Plane). Traditional manifold learning algorithms mainly reduce the dimension of especially high dimensional data points, where it is difficult to construct the polygon meshes. Note that all these works on applying Ricci flow to manifold learning are 2-dimensional, no matter whether they are discrete or continuous. They cannot be directly extended to higher dimensional cases.

\section{Basic Knowledge}
\label{sec:1}
Given data set $\{x_1,x_2,\cdots,x_N\subset \mathbb{R}^D\}$, where $N$ is the number of data points and $D$ their dimension. One fundamental assumption of traditional manifold learning is that $\{x_1,x_2,\cdots,x_N\subset \mathbb{R}^D\}$ lie on a $d$-dimensional Riemannian manifold $\left(\mathcal{M},g\right)$ embedded in the high dimensional Euclidean space $\mathbb{R}^D \left(d\ll D\right)$ , where $\mathcal{M}$ is the manifold itself, $g$ its Riemannian metric defined as the family of all inner products defined on all tangent spaces of $\mathcal{M}$ and $\mathbb{R}^D$ is called the ambient space of $\left(\mathcal{M},g\right)$. On each $T_p M$ the tangent space at point $p$, the Riemannian metric is a Euclidean inner product $g_p$. All the geometric intrinsic quantities (length, angle, area, volume, Riemannian curvature) of the Riemannian manifold can be computed with the Riemannian metric $g$.  
\subsection{Riemannian Curvature}
\label{sec:2}
In general, the Riemannian curvature tensor of a $d$-dimensional $\left(d\geq 3\right)$ Riemannian manifold is represented by a fourth-order tensor, which measures the curvedness of $\left(\mathcal{M},g\right)$ with respect to its ambient space $\mathbb{R}^D$. The directional derivative defined on a Riemannian manifold is Riemannian connection, represented by $\nabla$. Riemannian curvature tensor is a $\left(1,3\right)$-tensor defined by:
\begin{equation}
\mathcal{R}m\left(X,Y\right)Z = \nabla_X \nabla_Y Z- \nabla_Y\nabla_XZ-\nabla_{[X,Y]}Z,
\end{equation}
on vector fields $X,Y,Z$ \cite{26}. Using the Riemannian metric $g$, the Riemannian curvature tensor can be transformed to a $\left(0,4\right)$-tensor as follows:
\begin{equation}
\mathcal{R}m\left(X,Y,Z,W\right)=g\left(\mathcal{R}m(X,Y)Z,W\right).
\end{equation}
The trace of the Riemannian curvature tensor is a symmetric $\left(0,2\right)$-tensor called Ricci curvature tensor \cite{26}:
\begin{equation}
Ric\left(Y,Z\right) =tr \left(\mathcal{R}m\left(\cdot,X\right)Y\right).
\end{equation}
In differential geometry, the Riemannian metric is expressed by the first fundamental form. Regarding Riemannian sub-manifold, the Riemannian curvature tensor is captured by the second fundamental form $\mathcal{B}\left(X,Y\right)$ which is a bilinear and symmetric form defined on tangent vector fields $X,Y$. $\mathcal{B}\left(X,Y\right)$ is used to measure the difference between the intrinsic Riemannian connection $\nabla$ on M and the ambient Riemannian manifold connection $\widetilde{\nabla}$ on $\widetilde{\mathcal{M}}$, where $\mathcal{M}$ is embedded into $\widetilde{\mathcal{M}}$. The relationship between $\widetilde{\nabla}$ and $\nabla$ is shown by the following \textsl{Gauss formula} \cite{27}:
\begin{equation}
\widetilde{\nabla}_XY=\nabla_XY+\mathcal{B}\left(X,Y\right).
\end{equation}
The corresponding relationship between $\mathcal{R}m\left(X,Y,Z,W\right)$ of $\mathcal{M}$ and $\widetilde{\mathcal{R}m}\left(X,Y,Z,W\right)$ of $\widetilde{\mathcal{M}}$ is shown by the following \textsl{Gauss equation} \cite{27}:
\begin{equation}
\widetilde{\mathcal{R}m}\left(X,Y,Z,W\right)=\mathcal{R}m\left(X,Y,Z,W\right)-\langle\mathcal{B}\left(X,W\right),\mathcal{B}\left(Y,Z\right)\rangle+\langle\mathcal{B}\left(X,Z\right),\mathcal{B}\left(Y,W\right)\rangle.
\end{equation}
If the ambient space $\widetilde{\mathcal{M}}$ is an Euclidean space $\mathcal{R}^D$ then $\widetilde{\mathcal{R}m}\left(X,Y,Z,W\right)=0$. The Riemannian curvature of $\mathcal{M}$ can be fully captured by the second fundamental form:
\begin{equation}
\mathcal{R}m\left(X,Y,Z,W\right)=\langle\mathcal{B}\left(X,W\right),\mathcal{B}\left(Y,Z\right)\rangle-\langle\mathcal{B}\left(X,Z\right),\mathcal{B}\left(Y,W\right)\rangle.
\end{equation}
Under local coordinate system, the second fundamental form $\mathcal{B}$ can be represented by \cite{27}:
\begin{equation}
\mathcal{B}\left(X,Y\right)=\sum_{\alpha=d+1}^{D}h^\alpha\left(X,Y\right)\xi_\alpha,
\end{equation} 
where $\xi_\alpha, \left(\alpha = d+1,\cdots,D\right)$ is the normal vector field of $\mathcal{M}$ and $h^\alpha\left(X,Y\right)$ is shown by the second derivative of the embedding map about $X,Y$.

\subsection{Ricci Flow}
\label{sec:3}
Ricci flow is an intrinsic curvature flow on Riemannian manifold, which is the negative gradient flow of Ricci energy. The Ricci flow is defined by the following geometric evolution time dependent partial differential equation \cite{28}: $\frac{\partial g_{ij}}{\partial t}=-2Ric_{ij}$, where $g_{ij}=g\left(\partial_i,\partial_j\right)$. The Ricci curvature $Ric\left(g\right)$ can be considered as a Laplacian of the metric $g$, making Ricci flow equation a variation of the usual heat equation. A solution of a Ricci flow is a one-parameter family of metrics $g\left(t\right)$ on a smooth manifold $\mathcal{M}$, defined in a time interval $I\subset \mathbf{R}_+$. In the time interval $I \subset \mathbf{R}_+$ the Riemannian metric $g\left(t\right)$ satisfies the metric equivalence condition $e^{-2Ct} g\left(0\right)\leq g\left(t\right)\leq e^{2Ct} g\left(0\right)$ \cite{28}, where $|Ric|\leq C$ and $t\in I$. So the relative geodesic distance between arbitrary two neighborhood points on M is consistent under the Ricci flow. In global, the solution of Ricci flow only exists in a short period of time until the emergence of singular points. In \cite{29} researchers have worked out that the Riemannian manifold of dimension $d\geq 4$ can be transformed to a sphere under Ricci flow, when the sectional curvature $K$ satisfies $\frac{max K}{min K} <4$ everywhere.

\subsection{Statement of the Spherical Conditions}
\label{sec:4}
Now coming back to our problem, note that the above constructed patches on data points $x_i$ can be either elliptic with positive sectional curvature or hyperbolic with negative sectional curvature. In this paper we only consider the former case and use elliptic polynomial functions to evaluate the local patches on every point, so that the curvature operator at every point of the Riemannian manifold M is non-negative. In Ricci flow theory, when the intrinsic dimension of embedded manifold $d\leq 3$, the Riemannian manifold $\mathcal{M}$ with non-negative Ricci curvature can flow to a Sphere under Ricci flow. For $d\geq 4$, when the sectional curvature $K$ satisfies $\frac{max K}{min K} <4$ everywhere, the Riemannian manifold $\mathcal{M}$ with positive sectional curvature can flow to a Sphere under Ricci flow. So in the Ricci flow step of our algorithm, we apply Ricci flow process under the spherical conditions.
\section{Algorithm}
\label{sec:1}
In practice, it is difficult to analyze the global structure of a nonlinear manifold, especially when there is no observable explicit structure. Our key idea is that we can uncover the local structure of manifold and the global structure can be obtained by the alignment of the local structures. In this paper we decompose the embedded manifold into a set of overlapping patches and apply Ricci flow to these overlapping patches independently of each other to avoid singular points, since in general Ricci flow on global manifold may encounter singular points when blowing up. The global structure of deformed manifold under Ricci flow can be obtained from the deformed local patches with a suitable alignment.

\subsection{RF-ML algorithm}
In this subsection, we first give the algorithm process of RF-ML. Then we give some detailed analysis of these algorithm steps.\\
\\
\textbf{Our algorithm RF-ML is mainly divided into five steps:}

\begin{itemize}
	\item[1.] Find a local patch (neighborhood) $U_i,i=1,2,\cdots ,N$ for each data point $x_i,i=1,2,\cdots,N$ by using the $K$-nearest neighbor method. In order to find the $K$-nearest neighborhood, we need to define a distance metric to measure the closeness of arbitrary two points. The distance metric we use in this step is the Euclidean metric. 
	\item[2.] Construct a special local coordinate system on every point $x_i$. In the neighborhood $U_i$ we estimate the local patch information of $x_i$ with a covariance matrix $C_i$, $C_i=\sum_{x_k \in U_i}\left(x_k-\overline{x}_i\right)^T\left(x_k-\overline{x}_i\right)$, where $\overline{x}_i$ is the mean vector of the $K$-nearest data points. The first $d$ eigenvectors $\left(e_1,e_2,\cdots,e_d\right)$ with maximal eigenvalues of $C_i$ form a local orthonormal coordinate system of $x_i$ on $U_i$. The last $D-d$ eigenvectors $\left(e_{d+1},\cdots,e_D\right)$ form a local orthonormal coordinate system of normal space. 
	\item[3.] Determine the intrinsic dimension $d$ of local patches by the number of the $95\%$ principle components. In practice due to the different curvature of local patches, the local dimension $d_i$ of each patch $U_i$ may be in practice not all the same. We choose $d=max\{d_1,d_2,\cdots,d_N\}$.\\ 
	If $d=D$, stop the algorithm. The manifold's dimension is not reducible.\\
	If $d<D$, continue the algorithm.	
	\item[4.] Construct Ricci flow equations on local patches and then let the overlapping patches $U_i,i=1,\cdots,N$ flow independently into local spherical patches $Y_i,i=1,2,\cdots,N$ with constant positive Ricci curvature $C$. Details will be shown below. 
	\item[5.] Align the discrete sphere patches $Y_i,i=1,2,\cdots,N$ into a global subset $P$ of a sphere with positive curvature $C$, where $P\subset \mathbb{R}^D$. $p_i \in P$ is the corresponding representation of $x_i$. Details will be shown below. 
	\item[6.] Reduce the dimensionality of the subset $P$ using traditional manifold learning algorithms, where the distance metric between arbitrary two points on $P$ is the metric of the sphere other than the Euclidean. The $d$-dimensional representations are $\{z_1,z_2,\cdots,z_N\} \in \mathbb{R}^d$.	
\end{itemize}
The intuitive representation of algorithm RF-ML is shown in Figure 2. And the brief algorithm procedure is shown in Algorithm 1.\\
\\
\begin{figure}
	\centering
	\includegraphics[width=1.0\linewidth]{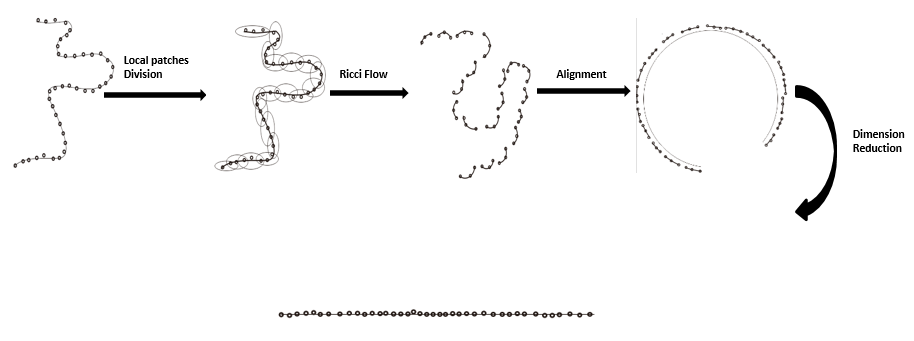}
	\caption{The intuitive process of algorithm RF-ML. The sub-images from left to right are respectively: input data points distributed on manifold M; the overlapping patches; overlapping patches flowing into discrete spherical patches using Ricci flow; alignment of spherical patches to a subset of a sphere. On the bottom is the representation of data points in low dimensional Euclidean space}
	\label{fig:figure2}
\end{figure}
\textbf{Step 4)} above is to minimize the value of the energy function such that the Ricci curvature at every point converges to a constant curvature. The curvature energy function is as follows:
\begin{equation}
E\left(Ric\right)=\int \|Ric-C\|^2 d\mathcal{M} \approx \sum_{i=1}^{N}|Ric\left(x_i\right)-C|^2,
\end{equation}
where $C$ is the value of constant non-negative Ricci curvature, $\mathcal{M}$ the embedded manifold where the $N$ data points are distributed on and $Ric$ is called the Ricci curvature. The convergence of Ricci flow proved in the following theorem 3.1 indicates that the energy function can provide the optimal solution. In order to obtain the minimum solution of the curvature energy function, we calculate the solution step by step with the help of Ricci flow. For the purposes of this paper, it is not our interest to learn the structure of a single Riemannian manifold $\left(\mathcal{M},g\right)$, but rather a one-parameter family of such manifolds $t\rightarrow \left(\mathcal{M}\left(t\right),g\left(t\right)\right)$, parameterized by a ‘time’ parameter $t$. The process is controlled by Ricci flow, until $\left(\mathcal{M}\left(t\right),g\left(t\right)\right)$ converges to a Riemannian manifold $\left(H,g\right)$ with constant curvature where the global set of patches $P$ (step 5) distributed on. In theorem 3.2 we prove that the Riemannian manifold $H$ is diffeomorphic to the original Riemannian manifold $\mathcal{M}$. 

The discrete Ricci flow equation as well as the corresponding iterative equations defined on the discrete data points $\{x_1,x_2,\cdots,x_N\}\in \mathbb{R}^D $ is constructed as follows:

Suppose the local coordinates of any point $x_j\in U_i$ under local ambient coordinates $\langle x_i;e_1,e_2,⋯,e_d,\cdots,e_D \rangle$ are represented as $\left(x_j^1,x_j^2,\cdots,x_j^D\right)$. The first $d$ coordinates can be seen as the local natural parameters of $U_i$. A smooth representation of local patch $U_i$ under the local coordinate system is described by:
\begin{equation}
f\left(x^1,x^2,\cdots,x^d\right)= ̇\left(x^1,\cdots,x^d,f^{d+1} \left(x^1,x^2,\cdots,x^d \right),\cdots,f^D \left(x^1,x^2,\cdots,x^d \right)\right),
\end{equation}
where $\left(x^1,\cdots,x^d \right)$ is a coordinate chart at $U_i$.

In this paper, we use \textsl{least square} method to approximate the analytic functions $f^\alpha,\alpha=d+1,\cdots,D$ under the local coordinate system constructed above. In order to guarantee the curvature operator on each patch being satisfied the spherical conditions as presented in section 2, we choose second-order elliptic polynomial functions to approximate the structures of local patches. The polynomial form of $f^\alpha$ is depicted as $f^\alpha \left(x\right)=W_\alpha \Phi^T,\alpha=d+1,\cdots,D$, where $W_\alpha=[a_0^\alpha,a_1^\alpha,\cdots,a_d^\alpha,a_{11}^\alpha,a_{12}^\alpha,\cdots,a_{\left(d,d\right)}^\alpha]$ is the coefficient vector to be determined by \textsl{least square} method and $\Phi=[1,x^1,\cdots,x^d,x^1 x^1,x^1 x^2,\cdots,x^d x^d]$ is the second-order polynomial basis vector.

Under the local smooth representation $f$ of $U_i$, the corresponding $d$ tangent vector basis at $x_i$ is given by $\{\frac{\partial f}{\partial x^1}\left(x_i\right),\frac{\partial f}{\partial x^2}\left(x_i\right),\cdots,\frac{\partial f}{\partial x^d}\left(x_i\right)\}$, where $\frac{\partial f}{\partial x^j}=\left(0,\cdots,1,\cdots,0,W_{d+1} \frac{\partial \Phi^T}{\partial x^j}\left(x_i\right),\cdots,W_D \frac{\partial \Phi^T}{\partial x^j}\left(x_i\right)\right)$. Then the local Riemannian metric tensor is shown as $G_i=[g_{jk}], g_{jk}=\langle \frac{\partial f}{\partial x^j}\left(x_i\right),\frac{\partial f}{\partial x^k}\left(x_i\right)\rangle$. The second fundamental form coefficient is shown as: $h_{jk}^\alpha\doteq \frac{\partial^2 f^\alpha}{\partial x^j \partial x^k}$. 

So the local Ricci flow equation defined on $x_i$ is represented as follows:
\begin{equation}
\frac{\partial}{\partial t}\left(\frac{\partial f}{\partial x^j}\cdot \left(\frac{\partial f}{\partial x^k}\right)^T\right)_{x_i} = -2 \sum_{\alpha=d+1}^{D}\sum_{l=1}^{d}\left(\frac{\partial^2 f^\alpha}{\partial x^l \partial x^l}\cdot \frac{\partial^2 f^\alpha}{\partial x^j \partial x^k}-\frac{\partial^2 f^\alpha}{\partial x^l \partial x^j}\cdot \frac{\partial^2 f^\alpha}{\partial x^l \partial x^k}\right)_{x_i}.
\end{equation}
In order to solve the Ricci flow equation, we need to discretize the differential operators on point clouds. Suppose $V_j=\left(\frac{\partial f^{d+1}}{\partial x^j},\frac{\partial f^{d+2}}{\partial x^j},\cdots,\frac{\partial f^D}{\partial x^j}\right)$, Eq.10 can be represented as the function of $V_j$, that is $\frac{\partial}{\partial t} V_j=F\left(V_j,\nabla V_j\right)$, where $F\left(V_j,\nabla V_j \right)=-\sum_{\alpha = d+1}^{D} \sum_{l=1}^{d}\left(\frac{\partial^2 f^\alpha}{\partial x^l \partial x^l}\cdot \frac{\partial^2 f^\alpha}{\partial x^j \partial x^j}-\frac{\partial^2 f^\alpha}{\partial x^l \partial x^j} \cdot \frac{\partial^2 f\alpha}{\partial x^l \partial x^j}\right)\cdot \left(\frac{\partial f}{\partial x^j}\right)^{T^+}.$ So the local Riemannian metric tensor as well as the corresponding Ricci curvature are iterated under the Ricci flow in the local neighborhood $U_i$ as follows: 
\begin{eqnarray}
V_j^{n+1}=V_j^n + \Delta t F\left(V_j^n,\nabla V_j^n\right), j=1,\cdots,d,
\\g_{jk}^{n+1}=\delta_{jk}+V_j^{n+1}\cdot {V_k^{n+1}}^T,
\\ \mathcal{R}m_{jk}^{n+1} = \sum_{l}\left(\nabla_l V_l^{n+1}\cdot \nabla_k{V_j^{n+1}}^T-\nabla_l V_j^{n+1}\cdot \nabla_l {V_k^{n+1}}^T\right).
\end{eqnarray}
Optimizing these update equations until $\mathcal{R}m_{jk}^T\rightarrow C$, where $C$ is a non-negative constant and $T$ is the number of total iterations. \\
\\
\textbf{In step 5) above}, after the Ricci flow converges at each $x_i$, the overlapping local patches $\{U_1,U_2,\cdots,U_N\}$ are flowing to a set of discrete local spherical patches $\{Y_1,Y_2,\cdots,Y_N\}$. We denote the global coordinates of $\{Y_1,Y_2,\cdots,Y_N \}$ as $\{P_1,P_2,\cdots,P_N\}$. The relationship between $\{Y_1,Y_2,\cdots,Y_N \}$ and $\{P_1,P_2,\\\cdots,P_N \}$ is linked by a set of global alignment maps, which shift the discrete local spherical patches to a global subset of a sphere. We want the global coordinates $p_{i_j} \in P_i$ satisfying the following equations, such that $p_{i_j}$ is obtained by the local affine transformation of $y_{i_j}$, where $y_{i_j}\in Y_{i}$. We have:
\begin{equation}
p_{i_j}=\bar{p}_i + Q_i y_{i_j} + \epsilon_j^{\left(i\right)}, i=1,\cdots, N, j=1,\cdots, K,
\end{equation}
where $N$ is the number of sample points, $K$ the nearest neighbor size parameter, $Q_i$ the unit of orthogonal transformation and $\epsilon_j^{\left(i\right)}$ the reconstruction error.

In order to obtain the optimized local affine transformation, we need to minimize the local reconstruction error matrix $E_i$:
\begin{equation}
min E_i = \sum_{i=1}^{N}\|P_i\left(I-\frac{1}{K}ee^T\right)-Q_iY_i\|_F^2,
\end{equation}
where $E_i = [\epsilon_1^{\left(i\right)},\epsilon_2^{\left(i\right)},\cdots,\epsilon_K^{\left(i\right)}]$.\\
Minimizing the above \textsl{least square error} is equal to solve the following eigenvalue problem:
\begin{equation}
B=SWW^TS^T,
\end{equation}
where $S=[S_1,S_2,\cdots,S_N]$, $PS_i=P_i$, $W=diag\left(W_{11},W_{22},\cdots, W_{NN}\right)$, $W_{ii}=\left(I-\frac{1}{K}ee^T\right)\left(I-Y_i^\dagger Y_i\right)$ and $Y_i^\dagger$ the generalized inverse matrix of $Y_i$.

Decomposing matrix $B$ using SVD method, we obtain $B=U\Lambda U^{-1}$, where the columns of $U$ are the unit orthogonal eigenvectors of $B$. $\Lambda$ is a diagonal matrix and the diagonal elements are the eigenvalues of $B$, which are arranged in ascending order. The optimal solution is given by the $D$ eigenvectors of the matrix $B$, corresponding to the $2nd$ to $\left(D+1\right)st$ smallest eigenvalues of $B$. However, the optimal data set $P$ that we finally need are distributed on a sphere with curvature $C$. So we need to give a set of constraints:
\begin{equation}
P_iP_i^T=\frac{1}{C}, i=1,2,\cdots,N.
\end{equation}
Under these constraints, we rewrite matrix $B$ as: $B=QRΛR^T Q^T$, where $R$ is a diagonal matrix, $R_{ii} = \frac{\sqrt[D]{C}}{\sqrt[2]{\|\left(U_{i1},U_{i2},\cdots, U_{iD}\right)\|}}$
$\left(i=2,\cdots,D+1\right)$, the rest $R_{jj}=1 \left(j\neq 2,\cdots,D+1\right)$.
The obtained $2nd$ to $\left(D+1\right)st$ columns of $Q$ are the optimal data set $P$, which are distributed on a sphere with curvature $C$.

\begin{algorithm}[tb]
	\caption{Applying Ricci flow to Manifold Learning (RF-ML)}
	\label{alg:example}
	\begin{algorithmic}
		\STATE {\bfseries Input:} Training data points $\{x_1,x_2,\cdots ,x_N\}\in \mathbb{R}^D$, neighbor size parameter $K$.\\
		\STATE {\bfseries Output:} Low dimensional representations $\{z_1,z_2,\cdots z_N\} \in \mathbb{R}^d$\\
		1.
		\begin{minipage}[t]{0.9\linewidth}
			\textbf{for} $i=1$ \textbf{to} $N$ \textbf{do} 
		\end{minipage}\\
		2.
		\begin{minipage}[t]{0.9\linewidth}
			\qquad Find $K$-nearest neighbors of $x_i$;
		\end{minipage}\\
		3.
		\begin{minipage}[t]{0.9\linewidth}
			\qquad Compute tangent space $T_{x_i} \mathcal{M}$.
		\end{minipage}\\	
		4.
		\begin{minipage}[t]{0.9\linewidth}
			\textbf{end for}
		\end{minipage}\\	
		5.
		\begin{minipage}[t]{0.9\linewidth}
			\textbf{Repeat}
		\end{minipage}\\
		6.
		\begin{minipage}[t]{0.9\linewidth}
			\qquad Update the Ricci flow equations using Eq.11, Eq.12, Eq.13.
		\end{minipage}\\
		7.
		\begin{minipage}[t]{0.9\linewidth}
			\textbf{Until} convergence
		\end{minipage}\\		 
		8.
		\begin{minipage}[t]{0.9\linewidth}
			Align the deformed $\{Y_1, Y_2,\cdots, Y_N\}$ to a complete subset $P$ of sphere.
		\end{minipage}\\		 
		9.
		\begin{minipage}[t]{0.9\linewidth}
			Use traditional manifold learning algorithms to reduce the dimension of spherical data points.
		\end{minipage}	  		
	\end{algorithmic}
\end{algorithm}

\subsection{Theoretical Analysis}

In this subsection, we give the theoretical anlysis of our algorithm. First we want to illustrate the convergence of Ricci flow, which has been proved by reaserchers. Second we give the relationship between the original manifold and the deformed space under Ricci flow.\\
\\
\textsl{\textbf{Theorem 3.1.} \cite{7} Assume that $\left(\mathcal{M},g_0\right)$ has weakly 1/4-pinched sectional curvatures in the sense that $0\leq K\left(\pi_1\right)\leq 4K\left(\pi_2\right)$ for all two-planes $\pi_1,\pi_2 \subset T_p \mathcal{M}$. Moreover, we assume that $\left(\mathcal{M}, g_0\right)$ is not locally symmetric. Then the normalized Ricci flow with initial metric $g_0$ exists for all time, and converges to a constant curvature metric as $t \rightarrow \infty$.}

This theorem has been proved by S. Brendle and R. Schoen \cite{7} in 2008. In global, the Ricci flow exists for all time, and converges to a constant curvature metric. For each local patch, the Ricci flow definitely exists and also converges to a postive constant curvature $C$.\\
\\
\textsl{\textbf{Theorem 3.2.} The global set of patches P (step 5) is distributed on a subspace $H$ of a sphere, where $H$ is a Riemannian manifold. In addition, $H$ is diffeomorphic to the original Riemannian manifold $M$.}\\
\\
\textbf{Proof:} The original data points $\{x_1,x_2,\cdots,x_N\}$ are distributed on $M$. At each data point $x_i$ there is a local neighborhood $U_i$, such that $\sum_{i} U_i$  can fully cover the manifold $M$. Thus $M$ has an enumerable set of basis. 

According to the local Ricci flow, at each data point $x_i$ there is a diffeomorphism $f_i$ between $U_i$ and $Y_i$. In the global alignment there is a diffeomorphism $g_i$ between $Y_i$ and $P_i$. Thus the map $g_i\circ f_i$ between $U_i$ and $P_i$ is also a local diffeomorphism. 
Let: 
\begin{equation}
l_i\left(p\right)=\left\{\begin{array}{l} 1, p \in U_i \\ 0, p \notin U_i\\
\end{array} 
\right.,
\end{equation}
\begin{equation}
o_i=\frac{l_i}{\sum_{j}l_j}, i=1,2,\cdots,N.
\end{equation}
Define:
\begin{equation}
o_i \cdot g_i \circ f_i = \left\{\begin{array}{l} o_i\cdot g_i \circ f_i \left(p\right), p \in U_i \\ 0, p \notin U_i\\
\end{array}
\right.,
i=1,2,\cdots,N.
\end{equation}
Let $f=\sum_{i} o_i\cdot g_i \circ f_i$. It is defined on the global manifold $\mathcal{M}$ and induces a local diffeomorphism in any local patch. Since $\mathcal{M}$ has an enumerable set of basis, according to the unit decomposition theorem of manifold, $f$ is a global diffeomorphism. Therefore $f\left(\mathcal{M}\right)$ is also a Riemannian manifold. It is obvious that the data set $P$ is distributed on the manifold $f\left(\mathcal{M}\right)$ which is just the manifold $H$ mentioned in the statement of this theorem. \\$\blacksquare$

\section{Experiments}

In this section we compare our algorithm RF-ML with other manifold learning algorithms on several synthetic datasets as well as four real world databases.

\subsection{Intrinsic Dimension}

One implicit fundamental assumption of traditional manifold learning is that the dimension of data sets at each point is all the same. But in practice the distribution of data points on a manifold may be not uniform, even the manifold's dimensions at different points may be not exactly the same. An intuitive example illustrating this situation is depicted in table 1. We list the numbers of neighborhoods with different dimensions in six groups of datasets: \textsl{ORL Face} \cite{33}, \textsl{Yale Face} \cite{35}, \textsl{Yale-B Face} \cite{35}, \textsl{Weizmann} \cite{34}, \textsl{Swiss Roll} \cite{32} and \textsl{Sphere} \cite{32}. We choose the neighborhood-size parameter $K=10$ and use PCA projection to obtain the 95\% principal components. The latter is approximately viewed as the dimension of the local neighborhoods and may not be the same for all the neighborhoods. Table 1 shows the number of neighborhoods in each dimension. 

\subsection{Dimensionality Reduction}

To evaluate the performance of our curvature-aware manifold learning algorithm RF-ML, we compare our method with several traditional manifold learning algorithms (PCA, Isomap, LLE, LEP, Diffu-Map, LTSA) on four sets of three dimensional data, including: \textsl{Swiss Roll}, \textsl{Sphere}, \textsl{Ellipsoid} and \textsl{Gaussian}. Swiss Roll is a locally flat surface, where the Gauss curvature (i.e. Ricci curvature of two dimensional manifolds) is zero everywhere. However, the Gauss curvatures of the other three datasets are not zero. The objective of this comparison is to map each data set to two dimensional Euclidean space and then to analyze the neighborhood preserving ratios (NPRs) \cite{22} of different algorithms. Table 2 shows the results of these comparisons with seven different algorithms on these four sets of data, where the K-nearest neighbor parameter is $K=10$. The neighborhood preserving ratio (NPR) \cite{22} is defined as follows: 

\begin{equation}
NB = \frac{1}{KN}\sum_{i=1}^{N} |\mathcal{N}\left(x_i\right) \bigcap \mathcal{N}\left(z_i\right)|,
\end{equation}
where $\mathcal{N}\left(x_i\right)$ is the set of $K$-nearest sample subscripts of $x_i$ and $\mathcal{N}\left(z_i\right)$ is the set of $K$-nearest sample subscripts of $z_i$. $|\cdot|$ represents the number of intersection points.
\begin{table*}[tbp]
	\centering  
	\caption{Numbers of neighborhoods in each dimension. The first row shows the six datasets. The second row shows the original dimension of each dataset. The rows from $3$ to $11$ respectively list the numbers of neighborhoods in dimension $1, 2, 3, 4, 5, 6, 7, 8, 9$. The 'Total' row shows the total number of data in each dataset. '$K$' denotes the neighborhood-size parameter, and 'Ratio' denotes the percentage of principal components. Among all the data in table 1, those labeled by '$*$' are from the original dataset and the other data comes from our experiment results.}
	\begin{tabular}{|l|c|c|c|c|c|c|}  
		\hline
		Databases &ORL Face &Yale Face &YaleB Face &Weizmann  &Swiss Roll &Sphere \\ \hline
		Org-dim &1024* &1024* &1024* &200* &3* &3* \\ \hline
		1 &0 &0 &0 &0 &0 &0 \\ \hline
		2 &0 & 0& 0 &90 &1000 &984 \\ \hline
		3 &0 &0 &38 &317 &  &16 \\ \hline
		4 &0 &0 &238 &366 & &\\ \hline
		5 &0 &0 &362 &315 & & \\ \hline
		6 &6 &0 &415 &251 & & \\ \hline
		7 &214 &9 &772 &349 & & \\ \hline
		8 &182 &156 &583 &387 & & \\ \hline
		9 & & &6 & & & \\ \hline
		Total &400* &165* &2414* &2075* &1000* &1000* \\ \hline
		$K$ &10 &10 &10 &10 &10 &10 \\ \hline
		Ratio &0.95 &0.95 &0.95 &0.95 &0.95 &0.95 \\ \hline
	\end{tabular}	
\end{table*} 
\begin{table*}[tbp]
	\centering  
	\caption{Neighborhood Preserving Ratio. In this experiment, we fix the neighborhood-size parameter $K = 10$.}
	\begin{tabular}{|l|c|c|c|c|c|c|c|}  
		\hline
		Methods &PCA &Isomap &LLE &LEP &Diffu-Map &LTSA &\textbf{RF-ML} \\ \hline
		Swiss Roll &0.5137 &0.8594 &0.6187 &0.3981 &0.4403 &0.6121 &\textbf{0.8594}\\ \hline
		Ellipsoid &0.4399 &0.6914 &0.6205 &0.7506 &0.5965 &0.4390 &\textbf{0.8702}\\ \hline
		Sphere &0.4815 & 0.6467& 0.5213 &0.7720 &0.5010 &0.5465 &\textbf{0.8684}\\ \hline
		Gaussian &0.9969 &0.9261 &0.9359 &6406 &0.9848 &\textbf{0.9970} &0.9909\\ \hline
	\end{tabular}	
\end{table*}

\begin{figure}
	\centering
	\includegraphics[width=0.7\linewidth]{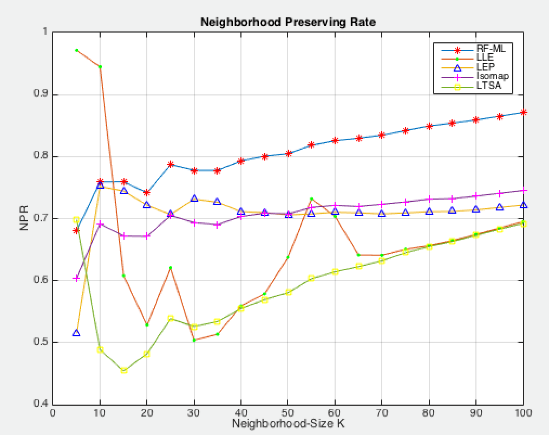}
	\caption{The neighborhood preserving rate of two-dimensional Ellipsoid embedded in 3-dimensional Euclidean space. Compute the NPRs under different neighborhood size parameter values $K$ with five manifold learning algorithms.}
	\label{fig:NPR}
\end{figure}

NPR measures the local structure preserving rate of the dimension reduction algorithms. Table 2 shows that for all but one dataset, the NPR of RF-ML is the best one among all algorithms. The Swiss Roll is a locally flat two-dimensional manifold. Its data structure is unchanged under Ricci flow. Therefore no Ricci flow is needed. As for the Sphere dataset, its Gauss curvature is a unique constant everywhere. There is no need of Ricci flow, too. Regarding Ellipsoid dataset and Gaussian dataset, note that the Gauss curvatures at different points are not always the same. When using RF-ML to reduce their dimensions, the NPR of Ellipsoid dataset is 87.95\%, 34.28\%, 66.58\%, 12.49\%, 73.33\%, 58.90\% respectively better over the other six traditional manifold learning algorithms. For Gaussian database, due to the characteristics of the dataset distribution the NPRs of PCA and LTSA are quite high. Compared with the six algorithms, the NPR of our method is not bad and also quite high. This clearly demonstrates that our curvature-aware algorithm RF-ML is more stable and the Ricci flow process preserves better the local structure of data points. 

Here we see another advantage of our algorithm. RF-ML is relatively stable under different values of neighbor-size parameter $K$. The 'stable' in this section means that the change of NPRs under different values of $K$ is relatively gentle. While traditional manifold learning algorithms are more sensitive to the neighbor-size parameter $K$. That is because those traditional algorithms implicitly assume the neighborhoods at each point be flat. Our method tries to find the intrinsic curvature structure of local patches at every point, so it is not sensitive to the neighbor-size parameter $K$. In this experiment, we evaluate the performance of our RF-ML algorithm compared with the other four algorithms (LLE, LEP, Isomap, LTSA) under different  values of neighbor-size parameter $K$ on the Ellipsoid dataset. The comparison results are showed in Figure 2. As the neighborhood-size parameter $K$ values are increasing, the NPR of our algorithm RF-ML grows continuously. But the NPRs of the other four manifold learning algorithms are unstable and especially sensitive to the different values of neighbor-size parameter $K$. 

\subsection{Real World Databases}

In this subsection, we give our experiments on four real world databases: ORL database \cite{33}, Yale Face database \cite{35}, Extended Yale Face B database \cite{35}, and USPS database \cite{38}.

The ORL Face database contains $10$ different images of each of $40$ distinct subjects. So this database totally contains $400$ images. For some subjects, the images were taken at different times, varying the lighting, facial expressions and facial details. 

The Yale Face database contains $165$ grayscale images of $15$ individuals. There are $11$ images per subject, one per different facial expression or configuration: center-light, glasses, happy, left-light, no glasses, normal, right-light, sad, sleepy, surprised, and wink.

The Extended Yale Face database B contains $5760$ single images of $10$ subjects. There are $576$ images per subject. For every subject in a particular pose, an image with ambient illumination was also captured.

The USPS database refers to numeric data obtained from the scanning of handwritten digits from envelopes by the U.S. Postal Service. There are $10$ classes of these handwritten digits from $0$ to $9$. The total number of this database is $9298$. The images here have been size normalized, resulting in $16 \times 16$ grayscale images. In our experiment, we respectively choose $700$ images for each digital class.

\begin{figure*} \centering    
	\subfigure[ORL database] { \label{fig:a}     
		\includegraphics[width=0.45\columnwidth]{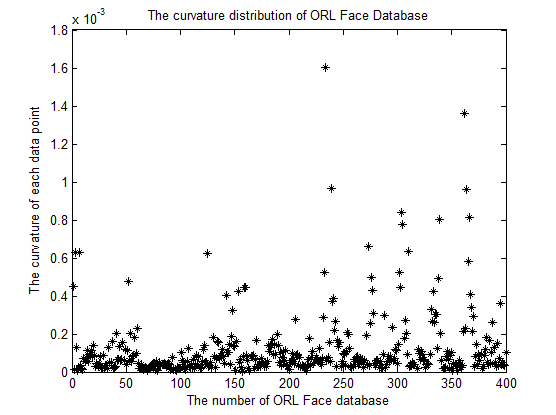}  
	}     
	\subfigure[Yale database] { \label{fig:b}     
		\includegraphics[width=0.45\columnwidth]{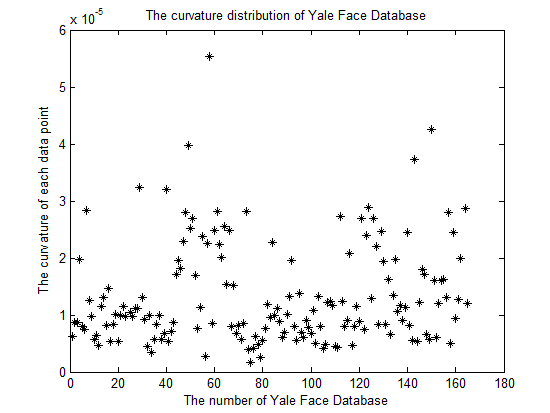}     
	}  
	\subfigure[YaleB database] { \label{fig:c}     
		\includegraphics[width=0.45\columnwidth]{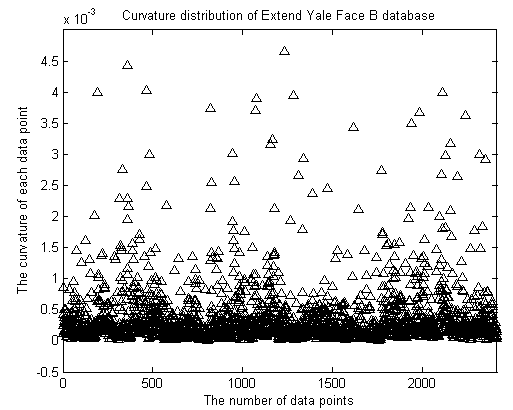}  
	}     
	\subfigure[USPS database] { \label{fig:d}     
		\includegraphics[width=0.45\columnwidth]{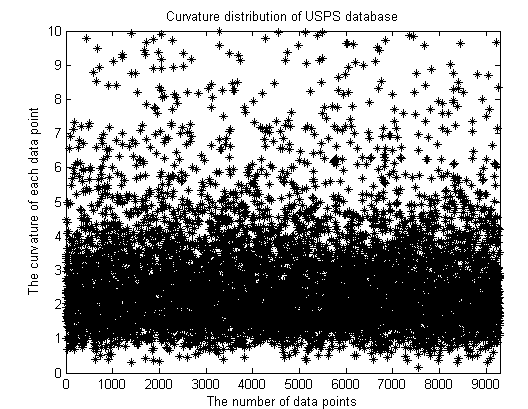}     
	}     
	\caption{The curvature distributions of ORL face database (a), Yale face database (b), YaleB face database (c), and USPS database (d).}     
	\label{fig}     
\end{figure*}

Our algorithm aims to regularize the curvature distribution of each database using Ricci flow. So we first analyze the curvature distributions of these four databases. The intuitive results are shown in Figure 3. From these four subfigures we can see that the curvature distribution of USPS database is more variable than the other three databases. One reason is that USPS database has enough samples to guarantee the samples distribution dense enough. The other reason is that the data points among different classes vary largely. For the other three databases, the curvature distributions are ranged from $0$ to $0.01$. The total numbers of ORL database and Yale Face database are respectively $400$ and $165$. Their distributions in high dimensional Euclidean space are especially sparse. So the curvature on each data point is relatively small.

Second, we compare our algorithms with traditional manifold learning algorithms on these four databases to test their classification accuracies. In this experiment, we first respectively use traditional manifold learning algorithms and our RF-ML algorithm to reduce the dimensions of these four databases. Then we use \textsl{Nearest Neighbor Classifier} method to test the recognition accuracies of different algorithms on these four databases. In the classification step, for these four databases, we respectively choose half images of each distinct subject as the training subset and the rest images of each subject as the test subset.  The experimental results are shown in Table 3. From Table 3 we can see that our proposed method RF-ML mostly outperforms the other four traditional manifold learning algorithms. Our method analyzes the curvature information of these four databases and uses Ricci flow to regularize the curvature information. So compared with traditional manifold learning algorithms, we uncover the intrinsic curvature information of these databases and add the information into the dimension reduction process.

Finally, we further compare our algorithm RF-ML with traditional manifold learning algorithms on USPS database to test the classification accuracies under different low dimension $d$. Comparing with the other three real world databases, the curvature distribution of USPS database varies higher than the other databases. So we do sufficiently comparative experiments on USPS database to test its performance under different low dimensions $d$. The final experimental results are shown in Figure 4. From Figure 4 we can see that the classification accuracies of our algorithm mostly outperform other traditional manifold learning algorithms under the different low dimension $d$. Overall, using Ricci flow to regularize the curvature distribution of dataset is a significant improvement of manifold learning.

\begin{table*}[tbp]
	\centering  
	\caption{Neighborhood Preserving Ratio. In this experiment, we fix the neighborhood-size parameter $K = 10$. In the dimension reduction step, we respectively choose $d=6,8,7,10$ as the intrinsic dimensions of ORL DB, Yale DB, YaleB DB, and USPS DB. In the recognition step, we respectively choose half of each dataset as the training dataset, the rest as the testing dataset.}
	\begin{tabular}{|l|c|c|c|c|c|c|c|}  
		\hline
		Methods &PCA &LPP &LLE &LEP &RF-ML \\ \hline
		ORL DB &$52.50 \pm 1.2$ &$62.50 \pm 2.1$ &$61.87 \pm 1.6$ &$63.82\pm 2.3$ &$65.04 \pm 1.3$ \\ \hline
		Yale DB &$35.73 \pm 1.4$ &$71.24 \pm 1.8$ &$69.05 \pm 2.2$ &$72.06\pm 1.5$ &$73.28 \pm 1.6$ \\ \hline
		YaleB DB &$63.72\pm 1.7$ & $67.26\pm 1.2$& $60.46 \pm 1.7$ &$72.62\pm 1.8$ &$73.95 \pm 2.1$\\ \hline
		USPS DB &$86.69\pm 0.9$ &$91.61\pm 1.3$ &$84.62\pm 1.5$ &$92.53\pm 1.1$ &$93.01\pm 1.7$ \\ \hline
	\end{tabular}	
\end{table*} 

\begin{figure}
	\centering
	\includegraphics[width=0.7\linewidth]{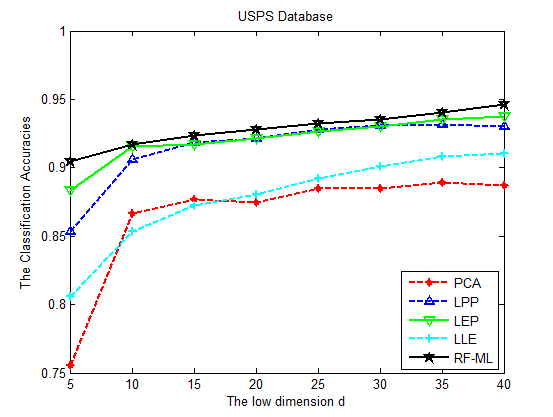}
	\caption{The classification accuracies of different manifold learning algorithms under different low dimensions $d$.}
	\label{fig:USPSFigure}
\end{figure}

\section{Conclusions and Future Work}
In the field of image recognition, to precisely describe the continuously changing images, one critical step is to assume the image set distributed on a lower dimensional manifold, which is embedded in the high dimension pixel space. Then one uses mathematical knowledge of manifold to deal with these datasets, such as dimensionality reduction, classification, clustering, recognition and so on. Manifold learning is an effective way to link the classical geometry with machine learning. Whether the manifold structure is uncovered exactly or not impacts the learning results directly. Many traditional manifold learning algorithms do not differentiate the varying curvature at different points of the manifold. Our method aims to excavate the power of Ricci flow and to use it to dynamically deform the local curvature to make the manifold's curvature uniform. The extensive experiments have shown that our method is more stable compared with other traditional manifold learning algorithms. There have been already papers published in the literature regarding applying Ricci flow to manifold learning as we have mentioned above. But to the best of our knowledge this paper is the first try to apply Ricci flow to high dimensional (dimension unlimited) data points for dimensionality reduction. One limitation of our algorithm is that RF-ML works only for manifolds with non-negative curvature. We will discuss the applicability of RF-ML algorithm to manifolds with negative Ricci curvature manifold in our next paper. Generally we believe that there will be lots of Ricci flow applications in manifold learning. 

\section*{Acknowledgements}

This work is supported by National Key Research and Development Program of China under grant 2016YFB1000902; National Natural Science Foundation of China project No.61232015, No.61472412, No.61621003; Beijing Science and Technology Project: Machine Learning based Stomatology; Tsinghua-Tencent- AMSS –Joint Project: WWW Knowledge Structure and its Application.

%
%




\end{document}